\documentclass[lettersize,journal]{IEEEtran}
\usepackage{amsmath,amsfonts}
\usepackage{algorithmic}
\usepackage{algorithm}
\usepackage{array}
\usepackage[caption=false,font=normalsize,labelfont=sf,textfont=sf]{subfig}
\usepackage{textcomp}
\usepackage{stfloats}
\usepackage{url}
\usepackage{verbatim}
\usepackage{graphicx}
\usepackage{cite}
\usepackage{times}
\usepackage{epsfig}
\usepackage{amsmath}
\usepackage{amssymb}
\usepackage{bm}
\usepackage{multirow}
\usepackage{balance}
\usepackage{color}
\usepackage{amsthm}
\newtheorem{definition}{Definition}

\begin{document}

\title{RFPose-OT: RF-Based 3D Human Pose Estimation \\via Optimal Transport Theory}

\author{Cong~Yu,
	Dongheng~Zhang,
	Zhi~Wu,
	Zhi~Lu,
	Chunyang~Xie,
	Yang~Hu,\\
	and~Yan~Chen*,~\IEEEmembership{Senior Member,~IEEE}
	% <-this % stops a space
	\thanks{
	% Copyright (c) 20xx IEEE. Personal use of this material is permitted. However, permission to use this material for any other purposes must be obtained from the IEEE by sending a request to pubs-permissions@ieee.org.
	*Corresponding author: Yan Chen (eecyan@ustc.edu.cn). 	
	% This work was supported by the National Natural Science Foundation of China under Grant 62172381.
	}
	\thanks{Cong Yu, Zhi Lu, and Chunyang Xie are with the School of Information and Communication Engineering, University of Electronic Science and Technology of China, Chengdu 611731, China. E-mail: congyu@std.uestc.edu.cn,
	zhilu@std.uestc.edu.cn, chunyangxie@std.uestc.edu.cn}% <-this % stops a space
	\thanks{Yang Hu is with the School of Information Science and Technology, University of Science and Technology of China, Hefei 230026, China. E-mail: eeyhu@ustc.edu.cn}
	\thanks{Dongheng Zhang, Zhi Wu, and Yan Chen are with the School of Cyber Science and Technology, University of Science and Technology of China, Hefei 230026, China. E-mail: dongheng@ustc.edu.cn, wzwyyx@mail.ustc.edu.cn, eecyan@ustc.edu.cn}}
% <-this % stops a space
\maketitle

\begin{abstract}
This paper introduces a novel framework, i.e., RFPose-OT, to enable the 3D human pose estimation from Radio Frequency (RF) signals. Different from existing methods that predict human poses from RF signals on the signal level directly, we consider the structure difference between the RF signals and the human poses, propose to transform the RF signals to the pose domain on the feature level based on Optimal Transport (OT) theory, and generate human poses from the transformed features. To evaluate RFPose-OT, we build a radio system and a multi-view camera system to acquire the RF signal data and the ground-truth human poses. The experimental results in basic indoor environment, occlusion indoor environment, and outdoor environment, all demonstrate that RFPose-OT can predict 3D human poses with higher precision than the state-of-the-art methods.
\end{abstract}

\begin{IEEEkeywords}
RF Sensing, Human Pose Estimation, Optimal Transport, Deep Learning
\end{IEEEkeywords}

\section{Introduction}
Due to the non-contact and privacy-preserving characteristics of radio signals, RF-based human sensing tasks have drawn increasing attention in recent years. Existing signal processing-based wireless sensing works mainly include human vital sign monitoring~\cite{conte2010ml, zhang2019breathtrack, yue2018extracting, zhang2019sj}, gesture recognition~\cite{niu2021understanding}, human gait authentication~\cite{ji2021one}, human position tracking~\cite{kotaru2015spotfi, chen2019residual, zhang2018multitarget, zhang2020mtrack, rampa2015physical, ito2020multi}, and human speed estimation~\cite{qian2018enabling, zhang2018wispeed}. With the development of deep learning, some learning-based methods~\cite{kim2018residual, chen2020speednet, li2021towards, zhang2021unsupervised, qiu2022radio} are proposed to handle wireless sensing tasks.

Besides the above classic wireless sensing tasks, some researchers~\cite{zhao2018through, li2019making, wang2019person, zhao2018rf, jiang2020towards,  yu2022rfgan, wu2022rfmask, song2022rf} explore utilizing RF signals to perceive human movements more finely based on deep learning methods, e.g., designing deep learning models to construct fine-grained human poses from radio signals. 
Specifically, \cite{zhao2018through} propose a teacher-student network model to estimate 2D human poses from FMCW signals. \cite{wang2019person} use a U-Net model to generate 2D human pose heatmaps from WiFi signals. Taking one step further, a 3D human pose estimation model based on RF signals is proposed in~\cite{zhao2018rf}, where the pose estimation task is regarded as a keypoint classification problem in the 3D space. WiFi-based 3D pose reconstruction has also been explored in~\cite{jiang2020towards}.

While achieving promising performance, existing RF-based human pose estimation methods transform RF signals to the human poses on the signal level directly, which ignores the structure difference between the RF signals and the human poses, i.e., the RF signals record human activities based on signal reflections that are processed as signal projection heatmaps, while target human poses are represented as skeletons in the real physical space based on human visual system. 
Therefore, in this paper, we pay more attention to the feature level, and propose RFPose-OT to transform the RF signals to the target pose feature domain based on OT theory, then generate pose keypoints from the transformed features. 
Specifically, three phases are designed to train RFPose-OT. 
1) We first train a pose encoder and a keypoint predictor to obtain the target human pose representations in the feature space with the supervision of ground-truth keypoints. 
2) Then, a RF encoder is trained to transform the RF signals to the target pose feature domain by utilizing the OT distance (defined through OT theory) as the training loss. 
3) Finally, we fine-tune the RF encoder and the keypoint predictor using ground-truth keypoints for fine-grained estimation results. 
Note that during RFPose-OT inference, the pose encoder can be thrown away, and only the RF encoder and the keypoint predictor are used for 3D human pose estimation. 

\begin{figure*}[t]
	\centering
	\includegraphics[width=0.85\textwidth]{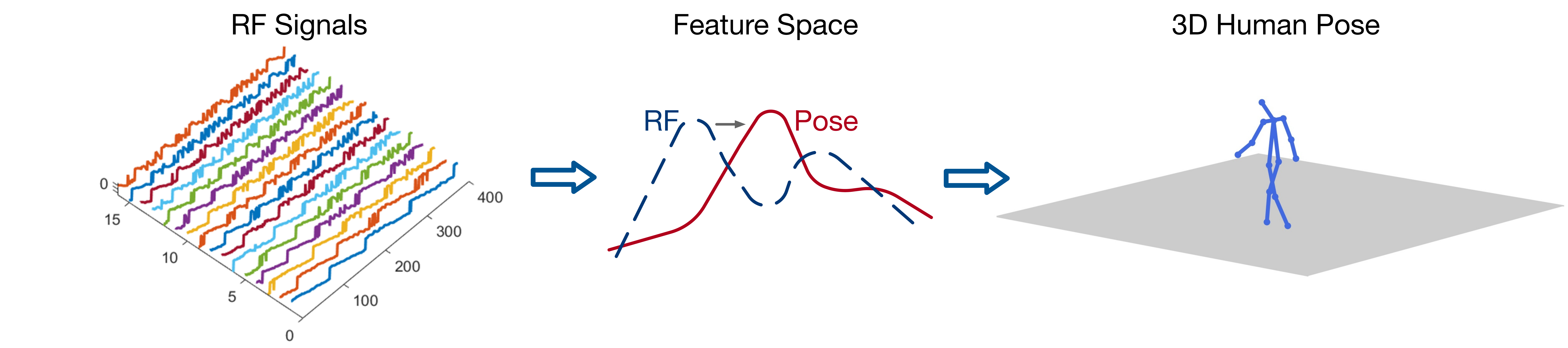}
	\caption{RFPose-OT transforms the RF signals to the pose domain to enable fine-grained 3D human pose estimation.}
	\label{fig:intror}
\end{figure*}

To evaluate RFPose-OT, we build a radio system to capture the horizontal and vertical RF signals and the signals are preprocessed to the RF heatmaps, the ground-truth human pose keypoints are obtained through a multi-view camera system. In both basic and occlusion indoor environments, we compare our proposed RFPose-OT with state-of-the-art methods and also conduct ablation studies. We further test RFPose-OT in an outdoor environment. The experimental results demonstrate that RFPose-OT can predict 3D human poses with high precision and outperform the alternative methods.

\section{Related Work}
With the popularity of radio devices, recent years have witnessed more and more interest in utilizing radio signals to enable sensing tasks~\cite{he2020wifi}.

\subsection{Classical Wireless Human Sensing} Based on signal processing algorithms, some researchers~\cite{patwari2013monitoring, zhang2019breathtrack} try to monitor human vital signs such as breathing by analyzing the heaving chest through radar or WiFi signals. 
% E.g., \cite{patwari2013monitoring} show that signal strength can be used to estimate the breathing rate and \cite{zhang2019breathtrack} design a breath tracking system via the phase variation of commodity WiFi.
In an indoor environment, some researchers focus on localization~\cite{majeed2015indoor} and tracking issues~\cite{wang2019witrace, zhang2018multitarget, zhang2020mtrack} and some others attempt to estimate human speed from radio signals~\cite{qian2018enabling, zhang2018wispeed}. 
Furthermore, some recent works also try to identify humans~\cite{zeng2016wiwho, hsu2019enabling} or recognize gestures~\cite{niu2021understanding} from radio signals based on signal processing technologies.

Meanwhile, deep learning methods have achieved remarkable achievements in many areas~\cite{lecun2015deep, zhang2018visual, yang2021multiple, li2018cyber,  ma2021associative, liu2022deep}. Therefore, besides utilizing signal processing methods, more and more researchers explore designing deep learning frameworks to push the limit of radio sensing tasks. 
For example, \cite{zhao2017learning} construct a conditional adversarial architecture to monitor sleep stages from radio signals via convolutional and recurrent neural networks.
\cite{xu2020leveraging} enable human breath detection from acoustic signals in noisy driving environments by training a deep learning model.
Taking a step further, deep neural networks can also learn electrocardiograms from millimeter wave signals~\cite{chen2022contactless}.
% Besides, \cite{fan2020learning} propose to re-identify persons by learning long-term representations, and radio-based gesture recognition is also explored in \cite{li2021towards} and \cite{zhang2021unsupervised} through the deep learning method.
In addition to the above coarse-grained human perception, much finer grained sensing tasks such as human pose estimation have also been explored recently.

\subsection{Human Pose Estimation}
Before utilizing radio signals to infer human poses, human pose estimation is a well-studied problem in computer vision literature~\cite{wei2016convolutional, cao2017realtime, fang2017rmpe, he2017mask, martinez2017simple, zheng20213d}.
However, vision-based human pose estimation methods often suffer from occlusion or bad illumination, while radio signals can traverse occlusion and do not rely on lights. Hence, radio-based human pose estimation has a wider range of application scenarios and has drawn increasing attention.
For example, \cite{zhao2018through} design a teacher-student network model to estimate 2D human poses from Frequency Modulated Continuous Wave (FMCW) signals with the supervision of a vision-based human pose estimation model, and they further extend the 2D version to the 3D version to achieve 3D human pose construction in~\cite{zhao2018rf}. After that, \cite{li2019making} use the FMCW signals to recognize human actions based on the human pose estimation results. 
Efforts have also been made to utilize WiFi signals to predict human poses, including the 2D~\cite{wang2019person} and the 3D~\cite{jiang2020towards} human pose estimation. RF-based human pose segmentation~\cite{wu2022rfmask} and visual synthesis~\cite{yu2022rfgan} are also explored recently.

However, the above existing methods usually follow the technologies in computer vision literature to predict human poses from radio signals directly, which ignores the structure difference between the radio signals and the human poses. Thus, in this paper, we propose to transform the radio signals to the pose feature domain first based on OT theory~\cite{monge1781memoire, kantorovich1942translocation}, and then predict human poses. 

\section{Primer of OT Theory}
\label{sec:primer}
In this paper, we utilize the OT distance to train model, which is defined through OT theory. OT theory discusses the problem of how to transport one distribution to another with the lowest cost, where the transport map is defined as follows:

\begin{definition}
	Map $T: \Omega \to \Psi$ transports measure $\mu \in \mathcal{P}(\Omega)$ to measure $\nu \in \mathcal{P}(\Psi)$, and we call $T$ a transport map, if for all $\nu$-measurable sets $B$,
	\begin{equation}
		\label{eqn:ot_tmap}
		\nu(B) = \mu(T^{-1}(B)). 
	\end{equation}
\end{definition}

\begin{figure}[h]
	\centering
	\includegraphics[width=0.96\columnwidth]{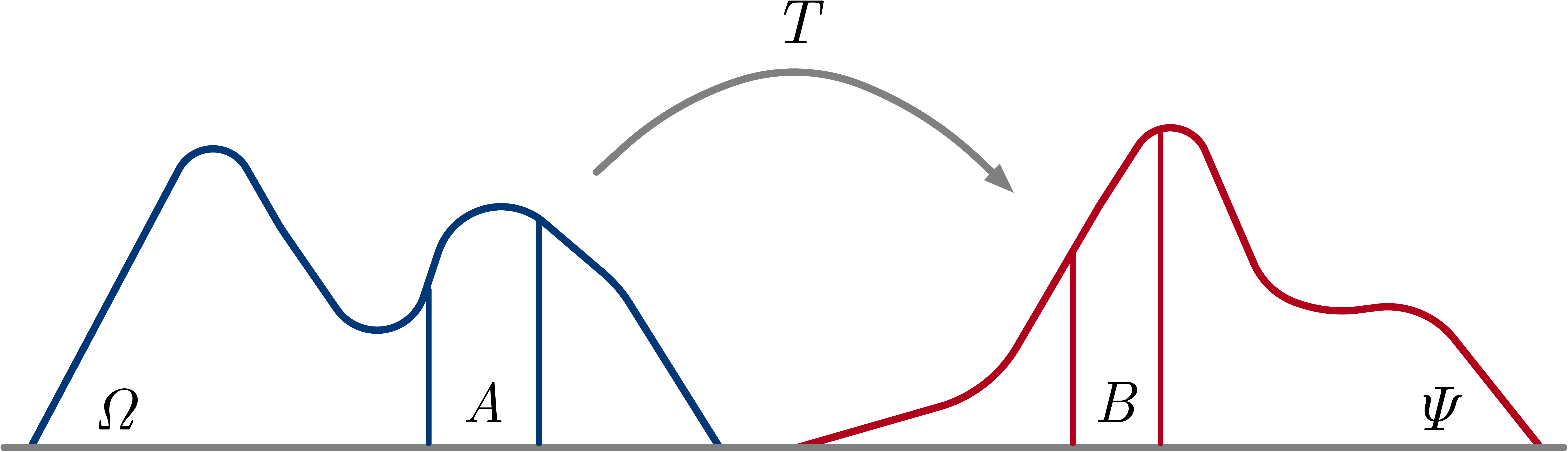}
	\caption{Transport map.}
	\label{fig:ot_map}
\end{figure}

Fig.~\ref{fig:ot_map} visualizes the transport map, where $A=\{z \in \Omega:T(z)\in B\}$ that $\mu(A)=\nu(B)$. Further, we define a cost function $C: \Omega \times \Psi \rightarrow [0, +\infty]$ to indicate the cost of transporting, and the overall transportation cost from $\mu$ to $\nu$ can be expressed using the Monge formulation~\cite{monge1781memoire} as 
\begin{equation}
	\label{eqn:monge}
	\mathcal{M}(T)=\int_{\Omega}C(z, T(z))\mathrm{d}\mu(z).
\end{equation}
OT theory is to find a transport map $T^\dagger$ to make the transportation cost $\mathcal{M}(T)$ minimum. Assume there exists the transport map $T^\dagger$, then the minimum cost $\mathcal{M}(T^\dagger)$ can be defined as the \textbf{OT distance} between $\mu$ and $\nu$ in the geometry space. If $\mathcal{M}(T^\dagger)=0$, we think $\Omega$ and $\Psi$ have the same distribution.

The generalization of Monge formulation is the Kantorovich formulation~\cite{kantorovich1942translocation}:
\begin{equation}
	\label{eqn:kantorovich}
	\mathcal{K}(\gamma)=\int_{\Omega \times \Psi}C(z, \hat{z})\mathrm{d}\gamma(z, \hat{z}),
\end{equation}
where $z \in \Omega$ and $\hat{z} \in \Psi$, $\gamma$ denotes the transport map from $\Omega$ to $\Psi$, which is subject to
\begin{equation}
	\begin{split}
		\label{eqn:kantorovich_st}
		&\gamma(A \times \Psi) = \mu(A)\\
		&\gamma(\Omega \times B) = \nu(B)
	\end{split}
\end{equation}
for all measurable sets $A \subseteq \Omega$, $B \subseteq \Psi$. Then, assume the optimal transport map $\gamma^\dagger$ exists, $\mathcal{K}(\gamma^\dagger)$ is the OT distance.

\section{RF Signal Collection and Preprocessing}
\begin{figure*}[t]
	\centering
	\includegraphics[width=0.95\textwidth]{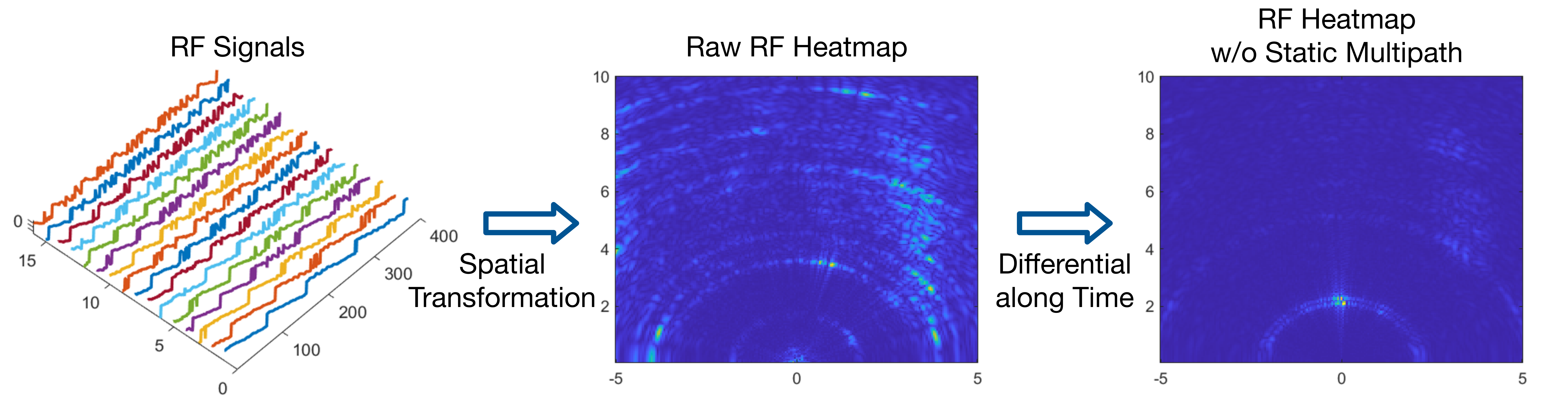}
	\caption{RF signal preprocessing. We first transform the collected RF signals to the spatial domain to get RF heatmaps, then apply the differential operation along time to remove static multipath.}
	\label{fig:rf_preprocess}
\end{figure*}

We utilize the millimeter wave (mmWave) radar with Multiple-In Multiple-Out (MIMO) antenna arrays to transceive RF signals. The received signals can be expressed as
\begin{equation}
	\label{eqn:r_signal}
	\bm{s}_{m,n}(t)=\sum_{l}a_l(t) \bm{\psi}_{l,m,n}(t)\bm{\phi}_{l,m,n}(t),
\end{equation}
where $m$, $n$, $l$, and $t$ denote the index of receiver antenna, frequency point, signal propagation path, and time, respectively, $a_l(t)$ is the complex attenuation coefficient, $\bm{\psi}_{l,m,n}(t)$ and $\bm{\phi}_{l,m,n}(t)$ are the phase shifts, which can be expressed as
% caused by Angle of Arrival (AoA) and Time of Flight (ToF):
\begin{equation}
	\label{eqn:r_phase1}
	\bm{\psi}_{l,m,n}(t)=e^{-j2\pi f_n\frac{(m-1)d\cos\theta_{l,m}(t)}{c}},
\end{equation}
\begin{equation}
	\label{eqn:r_phase2}
	\bm{\phi}_{l,m,n}(t)=e^{-j2\pi f_n\tau_{l,m}(t)},
\end{equation}
where $\theta_{l,m}(t)$ denotes the Angle of Arrival (AoA) and $\tau_{l,m}(t)$ denotes the Time of Flight (ToF), $d$ is the interelement distance of the antenna array, and $c$ is the signal propagation speed.

Since human poses are often defined in the rectangular coordinate system, thus, as shown in Fig.~\ref{fig:rf_preprocess}, we transform the received signals to the spatial domain:
\begin{equation}
	\label{eqn:rf2heatmap}
	\bm{S}_{x,y}(t)=\sum_{m}\sum_{n}\bm{s}_{m,n}(t)e^{j2\pi \varphi(x,y,m)},
\end{equation}
where $\bm{S}$ is the RF heatmap that represents the signals in the rectangular coordinate system, $x, y$ are the coordinates, and $e^{j2\pi \varphi(x,y,m)}$ is the phase shift determined by the spatial positions of the signal and the antenna.
According to the analysis in \cite{zhang2020mtrack}, there exists various static multipath in the raw RF heatmap. Therefore, we further apply the differential operation on $\bm{S}_{x,y}(t)$ along the time $t$ to remove the static multipath.

In this paper, we utilize two mmWave radars to obtain RF heatmaps on the horizontal and the vertical planes, respectively, which are denoted as $\bm{S}_H$ and $\bm{S}_V$ in the following sections.

\section{RFPose-OT}
In this paper, we propose a novel framework, i.e., RFPose-OT,  to enable 3D human pose estimation based on RF heatmaps $\bm{S}_H$ and $\bm{S}_V$. In the following, we first introduce the problem setup and explain the motivation of RFPose-OT. Then, we discuss the network structures and the model training of RFPose-OT.

\subsection{Problem Setup}
RFPose-OT aims to predict 3D human poses from the horizontal and vertical RF heatmaps. Since the human poses can be constructed by some body keypoints, the objective of RFPose-OT is to generate the 3D coordinates of the keypoints:
\begin{equation}
	\label{eqn:rfpose}
	\bm{\hat{p}}_{K\times3}=\text{RFPose-OT}(\bm{S}_H, \bm{S}_V),
\end{equation}
where $\bm{\hat{p}}_{K\times3}$ denotes the estimated keypoint coordinates, $K$ means the number of body keypoints and $3$ means the three dimensions of 3D space.

\begin{figure*}
	\begin{center}
		\includegraphics[width=0.85\textwidth]{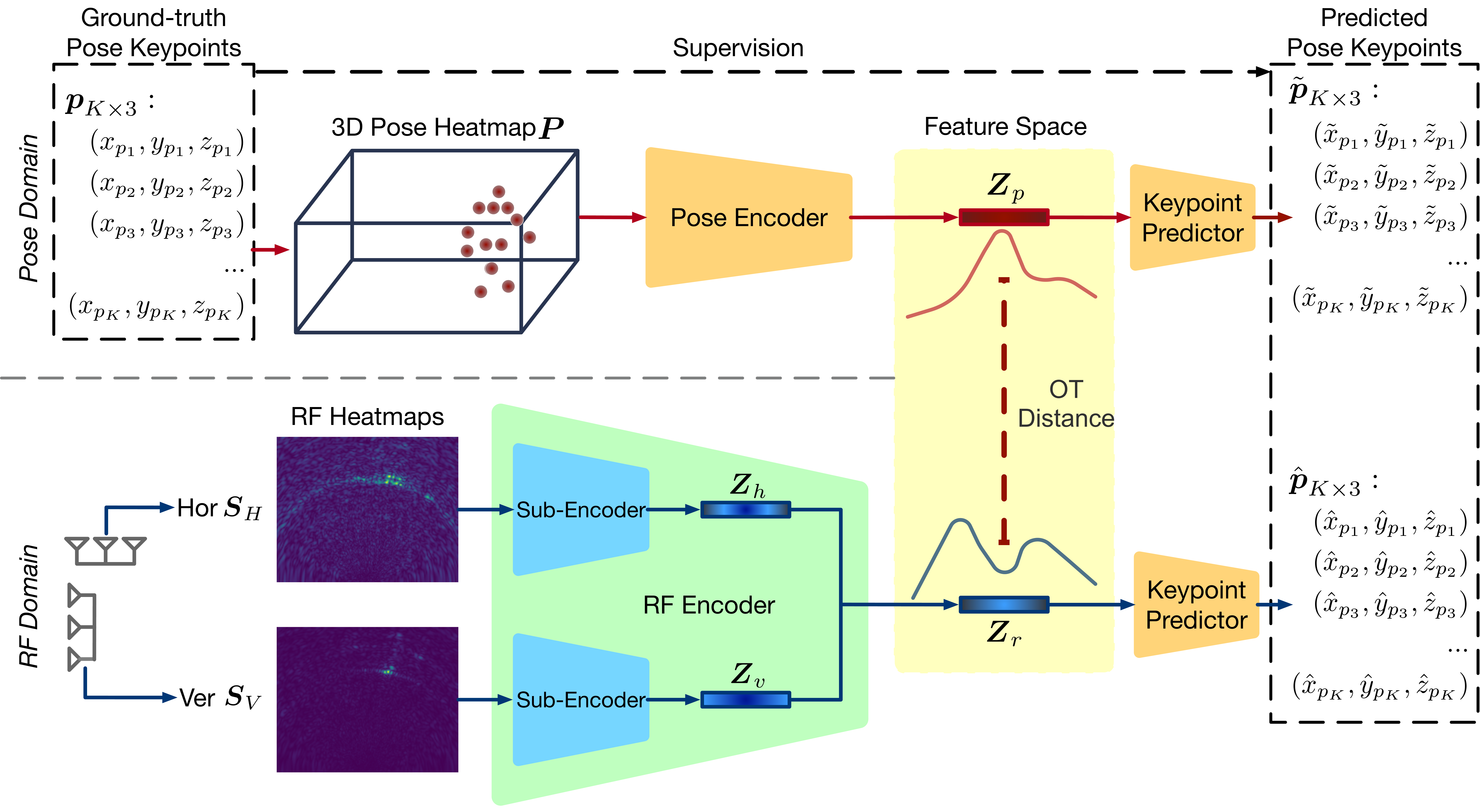}
	\end{center}
	\caption{The architecture of RFPose-OT. It consists of a pose encoder, a RF encoder, and a keypoint predictor. Once trained, only the RF encoder and the keypoint predictor are retained to predict 3D human poses from RF heatmaps.}
	\label{fig:model}
\end{figure*}

Although RF signals have been transformed to the spatial domain, from Fig.~\ref{fig:model}, we can see that $\bm{S}_H$ and $\bm{S}_V$ still have a totally different representation of human pose compared with the human pose skeleton that is based on the human visual system, i.e., the RF heatmaps and the human poses belong to two different feature domains, and predicting keypoints from RF heatmaps directly may be difficult. 

To tackle the above limitation, RFPose-OT tries to encode the RF heatmaps to the target pose domain.
Specifically, as shown in Fig.~\ref{fig:model}, we first learn the human pose embedding based on the ground-truth human poses, then train a RF encoder to transform the RF heatmaps to the human pose embedded space by using the OT distance as the training loss. Finally, the RF encoder and the keypoint predictor are fine-tuned to generate target human pose keypoints. Please note that once trained, the pose encoder can be thrown away and only the RF encoder and the keypoint predictor are needed in the inference phase.

\subsection{Pose Embedding}
To obtain the representation of human pose in the pose domain, 
as shown in Fig.~\ref{fig:model}, we draw a 3D pose heatmap based on the ground-truth pose keypoint coordinates. Specifically, for each keypoint, we synthesize a 3D point heatmap using the Gaussian kernel function as follows:
\begin{equation}
	\label{eqn:gaussian_3d}
	\bm{P}_{x,y,z}^{(k)}=e^{-\frac{(x-x_{p_k})^2+(y-y_{p_k})^2+(z-z_{p_k})^2}{2\sigma^2}},
\end{equation}
where $k$ denotes the index of keypoint, $x_{p_k}$, $y_{p_k}$ and $z_{p_k}$ denote the coordinates of the $k$-th keypoint on the X-Axis, Y-Axis and Z-Axis, respectively. After that, we combine all 3D point heatmaps together to get the 3D pose heatmap:
\begin{equation}
	\label{eqn:pose_heatmap}
	\bm{P}_{x,y,z}=\sum_{k=1}^{K}\bm{P}_{x,y,z}^{(k)},
\end{equation}
where $\bm{P}$ represents the human pose in 3D physical space.
Then, a pose encoder is followed to map the 3D pose heatmap $\bm{P}$ to the pose feature vector $\bm{Z}_p$:
\begin{equation}
	\label{eqn:Zp}
	\bm{Z}_p=E_P(\bm{P}),
\end{equation}
where $E_P$ denotes the pose encoder, $\bm{Z}_p$ is a manifold embedding code that contains the information of keypoint locations and the spatial relationship between the keypoints.
Finally, a keypoint predictor is designed to transform the pose feature vector $\bm{Z}_p$ to the pose keypoint coordinates:
\begin{equation}
	\label{eqn:kp_Zp}
	\bm{\tilde{p}}_{K\times3}=F(\bm{Z}_p).
\end{equation}

Obviously, the predicted keypoint coordinates $\bm{\tilde{p}}_{K\times3}$ from $\bm{P}$ should be the same as the ground-truth pose keypoint coordinates.
Hence, we train the pose encoder and the keypoint predictor using the following two measurements:
\begin{equation}
	\label{eqn:L_P}
	\mathcal{L}_{P}=\left \| \bm{\tilde{p}}_{K\times3}-\bm{p}_{K\times3} \right \|_{2},
\end{equation}
\begin{equation}
	\label{eqn:L_P0}
	\mathcal{L}_{PO}=\left \| (\bm{\tilde{p}}_{K\times3}-\frac{1}{K}\sum_{k}^{K}\tilde{p}_k)-(\bm{p}_{K\times3}-\frac{1}{K}\sum_{k}^{K}p_k) \right \|_{2},
\end{equation}
where $\tilde{p}_k$ denotes the $k$-th predicted keypoint coordinate and $p_k$ denotes the $k$-th ground-truth keypoint coordinate. Thus, $\frac{1}{K}\sum_{k}^{K}\tilde{p}_k$ and $\frac{1}{K}\sum_{k}^{K}p_k$ are the center point coordinates of the predicted pose and the ground-truth pose. Therefore, $\mathcal{L}_{P}$ measures the absolute location error of pose keypoints, and $\mathcal{L}_{PO}$ pays more attention to the relative human poses regardless of the absolute locations.

\subsection{Transport RF to Pose} 
After pose embedding, with the parameters of the pose encoder and the keypoint predictor fixed, we design a RF encoder to map the RF heatmaps to the RF feature vector. 
Specifically, as shown in Fig.~\ref{fig:model}, the RF encoder contains two sub-encoders to encode the horizontal and the vertical RF heatmap respectively, then, the extracted RF representations $\bm{Z}_h$ and $\bm{Z}_v$ are fused and mapped to the RF feature vector $\bm{Z}_r$:
\begin{equation}
	\label{eqn:Zr}
	\bm{Z}_r=E_R(\bm{S}_H, \bm{S}_V),
\end{equation}
where $E_R$ denotes the RF encoder, and $\bm{Z}_r$ has the same dimensions as the pose feature vector $\bm{Z}_p$.

Considering the geometry of the feature spaces, i.e., the RF feature space $\mathbb{Z}_r$ and the pose feature space $\mathbb{Z}_p$, we utilize the OT distance (recall the primer in Section~\ref{sec:primer}) to assess the divergence between them. Specifically, assume $\mu_r$ and $\mu_p$ are probability measures on spaces $\mathbb{Z}_r$ and $\mathbb{Z}_p$ respectively, and let $C: \mathbb{Z}_r \times \mathbb{Z}_p \rightarrow [0, +\infty]$ be a cost fuction where $C(\bm{Z}_r, \bm{Z}_p)$ measures the cost of transporting one unit of mass from $\bm{Z}_r \in \mathbb{Z}_r$ to $\bm{Z}_p \in \mathbb{Z}_p$, based on Kantorovich's OT theory, the OT distance can be expressed as 
\begin{equation}
	\label{eqn:L_OT}
	\mathcal{L}_{OT}=\int_{\mathbb{Z}_r \times \mathbb{Z}_p}C(\bm{Z}_r, \bm{Z}_p)\mathrm{d}\gamma(\bm{Z}_r, \bm{Z}_p)
\end{equation}
where $\gamma \in \mathcal{P}(\mathbb{Z}_r, \mathbb{Z}_p)$ denotes the optimal transport map, which indicates the amount of mass transported from $\bm{Z}_r$ to $\bm{Z}_p$ and satisfies the following marginal constraints 
\begin{equation}
	\begin{split}
		\label{eqn:st}
		&\gamma(\mathbb{Z}_r^i \times \mathbb{Z}_p) = \mu_r(\mathbb{Z}_r^i)\\
		&\gamma(\mathbb{Z}_r \times \mathbb{Z}_p^i) = \mu_p(\mathbb{Z}_p^i)
	\end{split}
\end{equation}
for all measurable sets $\mathbb{Z}_r^i \subseteq \mathbb{Z}_r, \mathbb{Z}_p^i \subseteq \mathbb{Z}_p$. 
The cost function $C(\bm{Z}_r, \bm{Z}_p)$ is set as $\left \| \bm{Z}_r - \bm{Z}_p \right \|_1$, then the optimal $\gamma(\bm{Z}_r, \bm{Z}_p)$ can be computed using the method in~\cite{bonneel2011displacement}. 
Ideally, we hope that the RF feature space $\mathbb{Z}_r$ has the same distribution as the pose feature space $\mathbb{Z}_p$ after RF encoder training, i.e., the OT distance is supposed to be $0$. Hence, we utilize $\mathcal{L}_{OT}$ as the objective function to train the RF encoder.

\begin{algorithm}[t]
	\caption{Training and fine-tuning algorithm of RFPose-OT.}
	\label{alg:rfpose_ot_algo}
	\begin{flushleft}
		\textbf{Set:}
		Batch size $b$, Learning rate $\eta$.\\
		\textbf{Initialize:} 
		Pose encoder parameters $\Phi_{E_P}$,
		RF encoder parameters $\Phi_{E_R}$,
		Keypoint predictor parameters $\Phi_{F}$.
	\end{flushleft}
	\begin{algorithmic}[1]
		\WHILE{$\Phi_{E_P}$ and $\Phi_{F}$ have not converged}
		\STATE Sample a batch of $\{\bm{p}_{K\times 3}\}$
		\STATE Update $\Phi_{E_P}, \Phi_{F}$ using Adam with:
		\STATE \quad
		$\Phi_{E_P} \leftarrow \Phi_{E_P} - \eta\frac{1}{b}\nabla_{\Phi_{E_P}}\sum_{i=1}^{b} (\mathcal{L}_{P}+\mathcal{L}_{PO})$
		\STATE \quad
		$\Phi_{F} \leftarrow \Phi_{F} - \eta\frac{1}{b}\nabla_{\Phi_{F}}\sum_{i=1}^{b} (\mathcal{L}_{P}+\mathcal{L}_{PO})$
		\ENDWHILE
		\WHILE{$\Phi_{E_R}$ has not converged}
		\STATE Sample a batch of $\{\bm{S}_H, \bm{S}_V, \bm{Z}_p\}$
		\STATE Update $\Phi_{E_R}$ using Adam with:
		\STATE \quad
		$\Phi_{E_R} \leftarrow \Phi_{E_R} - \eta\nabla_{\Phi_{E_R}}\mathcal{L}_{OT}$
		\ENDWHILE
		\WHILE{$\Phi_{E_R}$ and $\Phi_{F}$ have not converged}
		\STATE Sample a batch of $\{\bm{S}_H, \bm{S}_V, \bm{p}_{K\times 3}\}$
		\STATE Update $\Phi_{E_R}, \Phi_{F}$ using Adam with:
		\STATE \quad
		$\Phi_{E_R} \leftarrow \Phi_{E_R} - \eta\frac{1}{b}\nabla_{\Phi_{E_R}}\sum_{i=1}^{b} (\mathcal{L}_{R}+\mathcal{L}_{RO})$
		\STATE \quad
		$\Phi_{F} \leftarrow \Phi_{F} - \eta\frac{1}{b}\nabla_{\Phi_{F}}\sum_{i=1}^{b} (\mathcal{L}_{R}+\mathcal{L}_{RO})$
		\ENDWHILE
	\end{algorithmic}
\end{algorithm}

\subsection{Fine Tuning} 
After transporting the RF domain to the pose domain, we can estimate the pose keypoint coordinates from $\bm{Z}_r$ using the keypoint predictor: 
\begin{equation}
	\label{eqn:kp_Zr}
	\bm{\hat{p}}_{K\times3}=F(\bm{Z}_r).
\end{equation}

For better human pose estimation results, we further fine tune the RF encoder and the keypoint predictor, the objective functions are similar to Eqn.(\ref{eqn:L_P}) and Eqn.(\ref{eqn:L_P0}), where $\bm{\tilde{p}}_{K\times3}$ is replaced by $\bm{\hat{p}}_{K\times3}$:
\begin{equation}
	\label{eqn:L_R}
	\mathcal{L}_{R}=\left \| \bm{\hat{p}}_{K\times3}-\bm{p}_{K\times3} \right \|_{2},
\end{equation}
\begin{equation}
	\label{eqn:L_R0}
	\mathcal{L}_{RO}=\left \| (\bm{\hat{p}}_{K\times3}-\frac{1}{K}\sum_{k}^{K}\hat{p}_k)-(\bm{p}_{K\times3}-\frac{1}{K}\sum_{k}^{K}p_k) \right \|_{2}.
\end{equation}

The whole training and fine-tuning procedure is described in Algorithm~\ref{alg:rfpose_ot_algo}.

\begin{table}[t]
	\centering
	\caption{Pose encoder.}
	\label{tab:pose_encoder}
	\setlength{\tabcolsep}{8mm}
	\begin{tabular}{l|l}
		\hline
		Architecture & Parameters \\
		\hline
		\hline
		Conv-BN-ReLU & ks=5$\times$5, cs=32, st=2\\
		\hline
		Conv-BN-ReLU & ks=5$\times$5, cs=64, st=2\\
		\hline
		Conv-BN-ReLU & ks=5$\times$5, cs=64, st=1\\
		\hline
		Conv-BN-ReLU & ks=5$\times$5, cs=128, st=2\\
		\hline
		Conv-BN-ReLU & ks=5$\times$5, cs=128, st=1\\
		\hline
		Linear & 256\\
		\hline
	\end{tabular}
\end{table}

\begin{table}[t]
	\centering
	\caption{Sub-encoder in RF encoder.}
	\label{tab:rf_encoder}
	\setlength{\tabcolsep}{8mm}
	\begin{tabular}{l|l}
		\hline
		Architecture & Parameters \\
		\hline
		\hline
		Conv-BN-ReLU & ks=3$\times$3, cs=1, st=1\\
		\hline
		Conv-BN-ReLU & ks=5$\times$5, cs=8, st=2\\
		\hline
		Conv-BN-ReLU & ks=5$\times$5, cs=32, st=2\\
		\hline
		Conv-BN-ReLU & ks=5$\times$5, cs=128, st=2\\
		\hline
		Conv-BN-ReLU & ks=5$\times$5, cs=128, st=1\\
		\hline
		Conv-BN & ks=5$\times$5, cs=512, st=2\\
		\hline
	\end{tabular}
\end{table}

\begin{table}[t]
	\centering
	\caption{Keypoint predictor.}
	\label{tab:kp_pred}
	\setlength{\tabcolsep}{10mm}
	\begin{tabular}{l|c}
		\hline
		Architecture & Parameters \\
		\hline
		\hline
		Linear-LN-ReLU & 256\\
		\hline
		Linear-LN-ReLU & 256\\
		\hline
		Linear-LN-ReLU & 256\\
		\hline
		Linear & 14*3\\
		\hline
	\end{tabular}
\end{table}

\subsection{Inference Setting}
After training, the 3D pose heatmap and its embedding $\bm{Z}_p$, as well as the corresponding pose encoder, can be thrown away, and only the trained RF encoder is needed to estimate $\bm{Z}_r$ from the RF heatmaps. Since the RF encoder has been trained, it can directly map the RF heatmaps to the pose domain, then the output $\bm{Z}_r$ is input into the keypoint predictor to estimate the pose keypoint coordinates.

\section{Experiments}
In this section, we first describe the implementation details
and then conduct experiments to evaluate the performance of
the proposed RFPose-OT.

\subsection{Dataset} 
To collect RF signal data, we build a radio system using two mmWave radars (horizontal \& vertical), where each radar is equipped with 12 transmitters and 16 receivers with Multiple-In Multiple-Out (MIMO) antenna array. To avoid mutual interference, one radar works at 77GHz while the other works at 79GHz, both with 1.23GHz bandwidth. To obtain the ground-truth human poses, we build a multi-view camera system with 13 camera nodes, and a calibration method~\cite{zhang2000flexible} is applied to get the ground-truth 3D pose keypoint coordinates.

During data collection, we capture the RF signal reflections at 20Hz, and the camera system records videos at 10 FPS. The radio system and the camera system are synchronized using the network time protocol (NTP) through TCP connection, which achieves millisecond-level synchronization error. We collect the data under 10 indoor scenes. In total, we collect 89,090 RF signal samples and obtain the corresponding 3D pose keypoint coordinates, of which we use 80\% for training and the rest for testing. 

Since RF signals can traverse occlusions, we further collect more data in the occlusion environment as the additional testing set, where the radio system is occluded by baffles, then 1,180 RF signal samples are collected and the corresponding ground-truth poses can still be obtained by the camera system.

\subsection{Network Structure}
RFPose-OT is constructed by a pose encoder, a RF encoder, and a keypoint predictor.

The pose encoder consists of 5 convolutional layers and 1 linear layer. The RF encoder contains 2 sub-encoders with the same network structures, each sub-encoder consists of 6 convolutional layers. The keypoint predictor consists of 4 linear layers. More details of these network structures are shown in Table~\ref{tab:pose_encoder}, Table~\ref{tab:rf_encoder}, and Table~\ref{tab:kp_pred}, where Conv denotes the convolution layer, Linear denotes the linear layer, BN denotes the Batch Normalization, LN denotes the Layer Normalization, ReLU denotes the Rectified Linear Unit, ks denotes the kernel size, cs denotes the number of channels, st denotes the stride.

\subsection{Training Details} 
RFPose-OT is trained using Adam solver. The numbers of epochs for the pose embedding training and the RF transporting training are both 100, and the number of epochs for the fine-tuning is set to 50. The initial learning rate is set to 0.002 and decays by half every 10 epochs for all training phases. The batch size during model training is always 64. We implement our proposed RFPose-OT using PyTorch and all experiments can be run on a commodity workstation with a single GTX-1080 graphics card.

\begin{figure*}[t]
	\begin{center}
		\includegraphics[width=\textwidth]{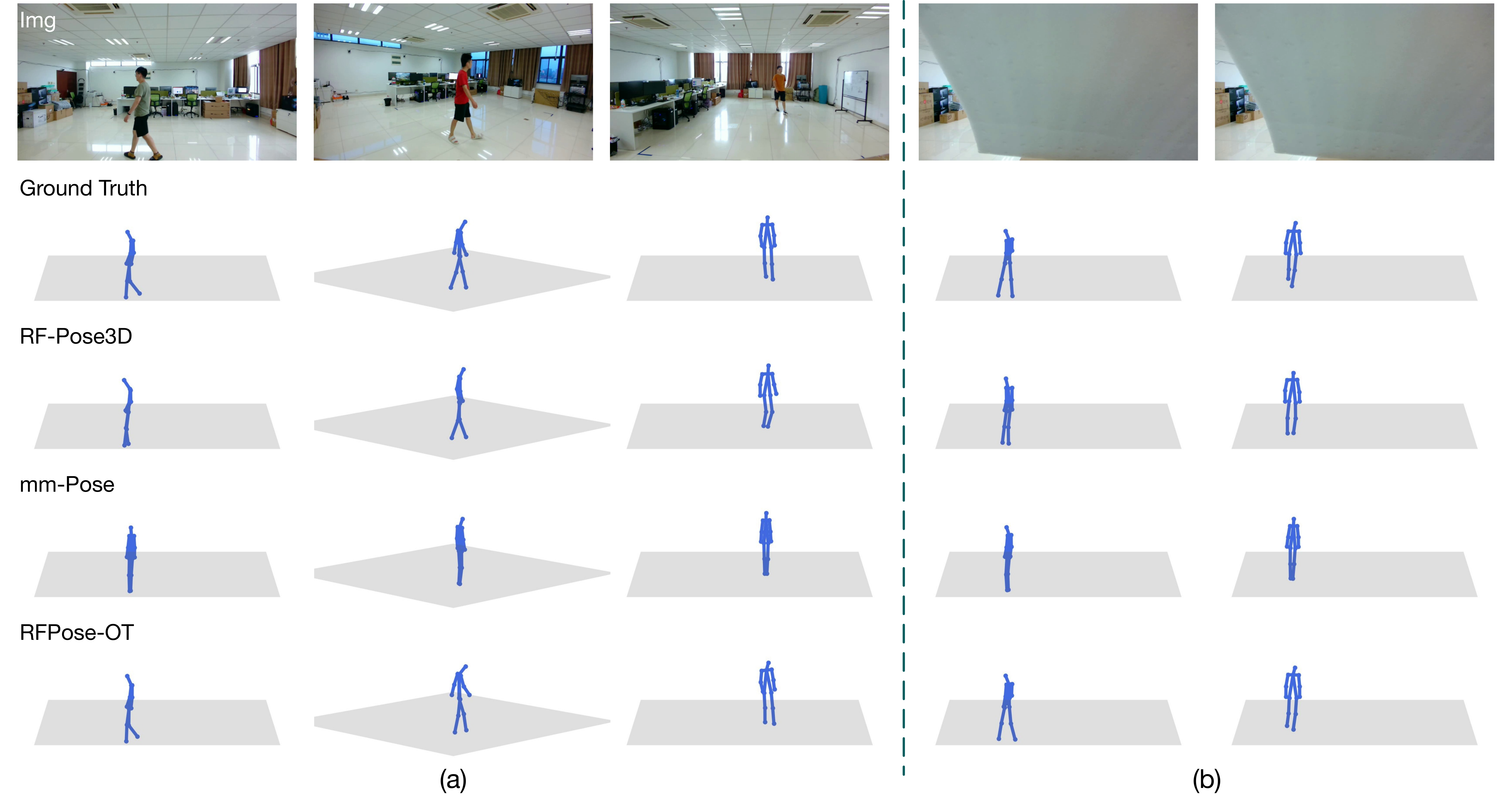}
	\end{center}
	\caption{Qualitative results of different methods in the (a) basic environment and (b) occlusion environment. 
		The 1st row shows the images captured by a camera that is attached to the radio system, the 2nd row shows the ground-truth 3D human poses, and the 3rd to 5th rows show the 3D human poses estimated by RF-Pose3D~\cite{zhao2018rf}, mm-Pose~\cite{sengupta2020mm}, and our proposed RFPose-OT, respectively.}
	\label{fig:results}
\end{figure*}

\begin{table*}[t]
	\centering
	\caption{Quantitative evaluation results of different methods in the (a) basic environment and (b) occlusion environment. (unit: cm)}
	\label{tab:baseline}
	\setlength{\tabcolsep}{3.6mm}
	\begin{tabular}{l|l|cccccccc|c}
		\hline
		Envs & Methods & Nose & Neck & Shoulders & Elbows & Wrists & Hips & Knees & Ankles & Overall \\
		\hline
		\hline
		\multirow{3}{*}{(a)} & RF-Pose3D~\cite{zhao2018rf} & 8.11 & \textbf{5.21} & 7.57 & 9.92 & 15.74 & 6.64 & 11.31 & 21.10 & 11.27 \\
		& mm-Pose~\cite{sengupta2020mm}& 8.19 & 5.30 & 7.23 & 9.67 & 15.29 & \textbf{6.20} & 10.83 & 19.04 & 10.72 \\
		& RFPose-OT & \textbf{7.90} & 6.14 & \textbf{6.76} & \textbf{7.99} & \textbf{11.67} & 6.39 & \textbf{8.34} & \textbf{12.60} & \textbf{8.68} \\
		\hline
		\multirow{3}{*}{(b)} & RF-Pose3D~\cite{zhao2018rf} & \textbf{6.53} & 4.86 & 6.65 & 8.75 & 14.05 & 6.95 & 11.26 & 21.52 & 10.70 \\
		& mm-Pose~\cite{sengupta2020mm} & 6.64 & \textbf{3.88} & \textbf{6.34} & 9.16 & 14.84 & 6.98 & 11.28 & 19.28 & 10.45 \\
		& RFPose-OT & 7.85 & 6.42 & 6.78 & \textbf{7.90} & \textbf{11.41} & \textbf{6.82} & \textbf{9.35} & \textbf{14.05} & \textbf{9.07} \\
		\hline
	\end{tabular}
\end{table*}

\subsection{Metric} 
To assess the precision of human pose estimation, we calculate the Spatial Location Error (SLE) between the predicted keypoints and the corresponding ground-truth keypoints using Euclidean distance:
\begin{equation}
	\label{eqn:metric}
	\text{SLE}_k=\frac{1}{U}\sum_{u=1}^{U}\left\| \hat{p}_k^{(u)}-p_k^{(u)}\right\|_2,
\end{equation}
where $k$ denotes the index of keypoint, $u$ denotes the $u$-th test sample, and $U$ is the number of test samples. 

\begin{table*}[t]
	\centering
	\caption{Quantitative evaluation results of different ablation models and the full model in the (a) basic environment and (b) occlusion environment. (unit: cm)}
	\label{tab:ablation}
	\setlength{\tabcolsep}{3.0mm}
	\begin{tabular}{l|l|cccccccc|c}
		\hline
		Envs & Models & Nose & Neck & Shoulders & Elbows & Wrists & Hips & Knees & Ankles & Overall \\
		\hline
		\hline
		\multirow{3}{*}{(a)} & RFPose-OT w/o $\mathcal{L}_{OT}$ & 8.69 & 6.82 & 7.58 & 8.97 & 12.93 & 7.19 & 9.31 & 14.11 & 9.69 \\
		& RFPose-OT w/o $\mathcal{L}_{PO+RO}$ & 8.32 & \textbf{6.04} & 7.01 & 8.52 & 12.66 & \textbf{6.30} & 8.57 & 13.94 & 9.17 \\
		& RFPose-OT (full) & \textbf{7.90} & 6.14 & \textbf{6.76} & \textbf{7.99} & \textbf{11.67} & 6.39 & \textbf{8.34} & \textbf{12.60} & \textbf{8.68} \\
		\hline
		\multirow{3}{*}{(b)} & RFPose-OT w/o $\mathcal{L}_{OT}$ & 7.87 & 6.64 & 7.35 & 8.69 & 12.10 & 7.57 & 10.11 & 15.27 & 9.76 \\
		& RFPose-OT w/o $\mathcal{L}_{PO+RO}$ & \textbf{7.83} & \textbf{6.11} & 7.01 & 8.36 & 12.04 & \textbf{6.61} & 9.51 & 15.55 & 9.44 \\
		& RFPose-OT (full) & 7.85 & 6.42 & \textbf{6.78} & \textbf{7.90} & \textbf{11.41} & 6.82 & \textbf{9.35} & \textbf{14.05} & \textbf{9.07} \\
		\hline
	\end{tabular}
\end{table*}

\begin{figure*}[t]
	\centering
	\includegraphics[width=0.98\textwidth]{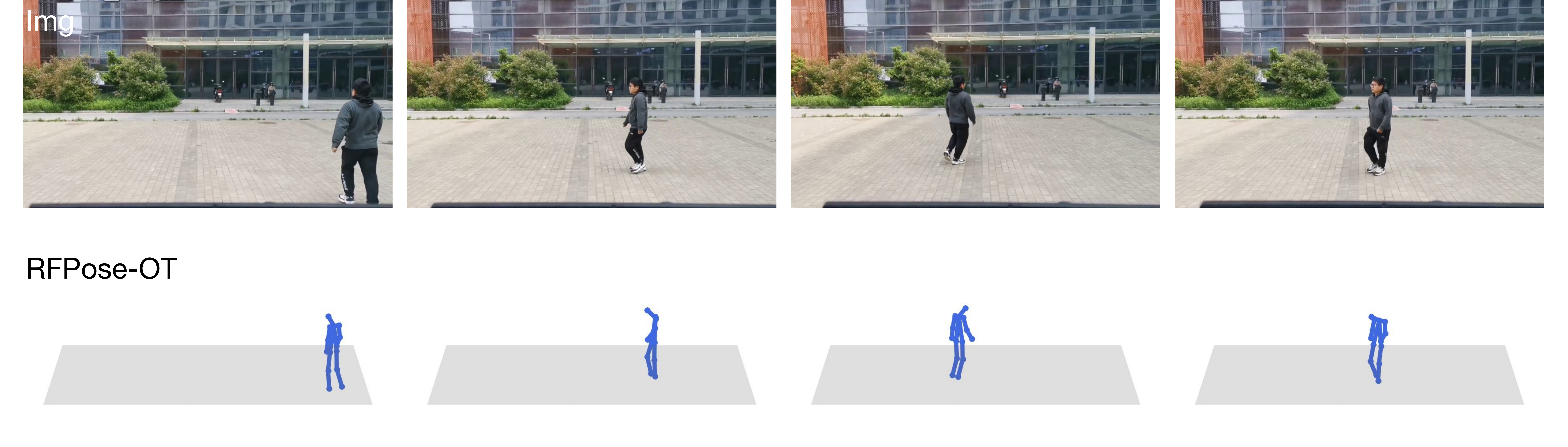}
	\caption{Qualitative results by RFPose-OT in an outdoor environment.}
	\label{fig:outdoor}
\end{figure*}

\begin{figure*}[t]
	\begin{center}
		\includegraphics[width=0.97\textwidth]{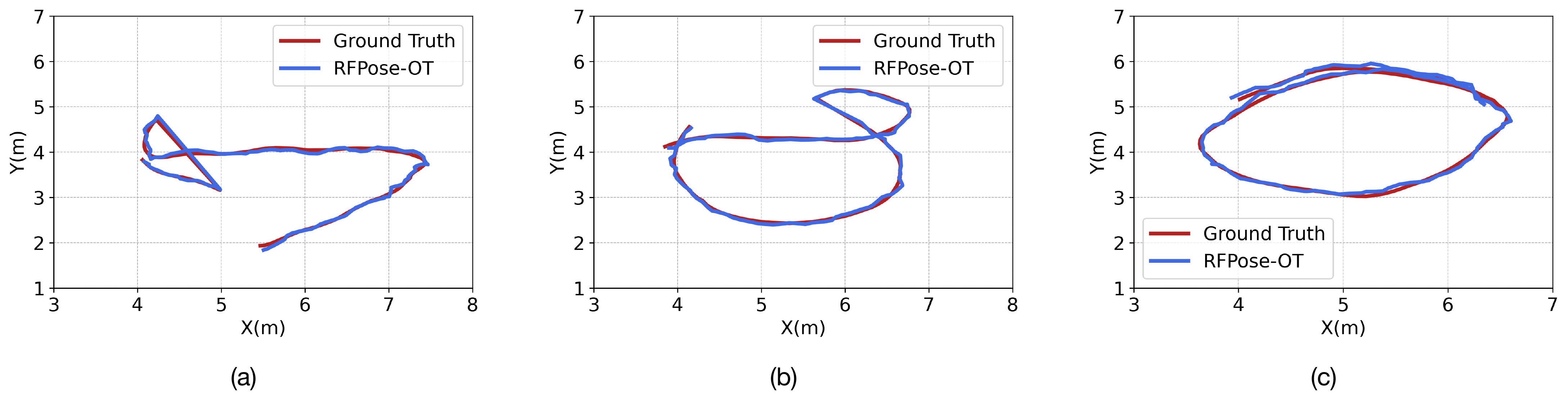}
	\end{center}
	\caption{Trajectories of a moving person in the (a)+(b) basic environment and (c) occlusion environment. The red line is the ground-truth trajectory and the blue line is the predicted trajectory.}
	\label{fig:tracking}
\end{figure*}

\subsection{Comparison with Baselines} 
We compare our proposed RFPose-OT with RF-Pose3D~\cite{zhao2018rf} and mm-Pose~\cite{sengupta2020mm}.

\begin{itemize}
	\item[1.] \textbf{RF-Pose3D:} RF-Pose3D is a classic deep learning model for 3D human pose estimation from RF signals, which regards pose estimation as a keypoint classification problem and is constructed by 18 convolution layers.
	\item[2.] \textbf{mm-Pose:} mm-Pose is a real-time 3D human pose estimation model based on mmWave radars, which consists of 6 convolution layers and 4 linear layers, and predicts the coordinates of human pose keypoints from the horizontal and the vertical RF signals directly.
\end{itemize}

The quantitative comparison results are summarized in Table~\ref{tab:baseline} (a), from which we can find that RFPose-OT outperforms baseline methods in almost all pose keypoints, especially gets much higher estimation precision in small body parts, e.g., wrists and ankles, which means the pose perception of our proposed RFPose-OT is much more fine-grained. We also show the qualitative results in Fig~\ref{fig:results} (a), from which we can see that although RF-Pose3D and mm-Pose can predict correct human locations, inaccurate poses are generated, while RFPose-OT can generate target 3D human poses that are consistent with the ground truth.

Compared with vision-based pose estimation method, RF-based pose estimation model can work in the occlusion scene. Thus, we further test RFPose-OT and the baseline methods in the additional testing set that is collected in the occlusion environment. The quantitative and qualitative results are shown in Table~\ref{tab:baseline} (b) and Fig.~\ref{fig:results} (b), from which we can see that RFPose-OT can still estimate 3D human poses from RF signals with high precision and outperforms alternative methods.
Above experimental results demonstrate the effectiveness of our proposed RFPose-OT model in the occlusion environment.

\subsection{Ablation Study} 
In this subsection, we further conduct ablation experiments to discuss the effects of some components in the RFPose-OT model.

\begin{itemize}
	\item[1.] \textbf{RFPose-OT w/o $\mathcal{L}_{OT}$:} In our full RFPose-OT model, an OT loss $\mathcal{L}_{OT}$ is designed for training the RF encoder to enable the transformation from the RF domain to the pose domain. The ablation model RFPose-OT w/o $\mathcal{L}_{OT}$ removes this component, i.e., RFPose-OT w/o $\mathcal{L}_{OT}$ predicts pose keypoints from the RF heatmaps directly.
	\item[2.] \textbf{RFPose-OT w/o $\mathcal{L}_{PO+RO}$:} In the pose embedding and the fine-tuning phases, $\mathcal{L}_{PO}$ and $\mathcal{L}_{RO}$ are proposed to pay more attention to the relative human poses. In RFPose-OT w/o $\mathcal{L}_{PO+RO}$, we discuss the effects of $\mathcal{L}_{PO}$ and $\mathcal{L}_{RO}$.
\end{itemize}

The quantitative evaluation results in the basic and the occlusion environments are shown in Table~\ref{tab:ablation}. We can find that our full RFPose-OT model outperforms the RFPose-OT w/o $\mathcal{L}_{OT}$ model in all keypoints, which means the OT loss $\mathcal{L}_{OT}$ we proposed in this paper is valuable, i.e., transporting the RF domain to the pose domain first and then predicting pose keypoints can improve the pose estimation precision. Besides, compared with RFPose-OT w/o $\mathcal{L}_{PO+RO}$, the full RFPose-OT model also performs better, which means $\mathcal{L}_{PO}$ and $\mathcal{L}_{RO}$ also contribute to the pose estimation.

\subsection{Performance in the Outdoor Scene}
Even though the RFPose-OT model is trained in the indoor environment, we still directly apply it to a new outdoor environment to evaluate its generalization. Since it is difficult to move the multi-view camera system to the new scenario to provide the ground truth poses, here we only qualitatively evaluate the performance. 
In Fig.~\ref{fig:outdoor}, we show the generated 3D poses by our model and the corresponding snapshots, from which we can see that our model can also correctly estimate 3D human poses from RF signals in the outdoor environment. This confirms the cross-environment generalization ability of RFPose-OT.

\section{Discussion}
\subsection{Model Complexity}
In this subsection, we calculate the amount of parameters (Params) and the amount of multiply-accumulate operations (MACs) to evaluate the complexity of our proposed RFPose-OT, and further test RFPose-OT on a single GTX-1080 graphics card to calculate the average running time for predicting one frame of human pose from RF signals. The results are shown in Table~\ref{tab:model_complexity}, from which we can see that RFPose-OT can support real-time processing.
\begin{table}[H]
	\centering
	\caption{Model complexity and running time.}
	\label{tab:model_complexity}
	\setlength{\tabcolsep}{6.5mm}
	\begin{tabular}{ccc}
		\hline
		Params & MACs & Time\\
		\hline
		\hline
		7.024 M & 1.302 G & 0.029 s/Frame \\
		\hline
	\end{tabular}
\end{table}

\subsection{Trajectory Tracking}
RFPose-OT is a fine-grained human activity sensing framework. Obviously, it can be used for handling some classic wireless sensing tasks, e.g., tracking the moving person. 
In this subsection, we utilize the trained RFPose-OT to track a subject who is asked to walk randomly in the basic and the occlusion scenes. The ground-truth trajectories are obtained by the multi-view camera system. RF signals are input into RFPose-OT, and we calculate the average value of the horizontal coordinates of the output keypoints as the predicted trajectories. 
The quantitative error are shown in Table~\ref{tab:trajectorie}, and
the qualitative results are shown in Figure~\ref{fig:tracking}, from which we can see that RFPose-OT can recover the moving trajectories in both basic and occlusion environments.
\begin{table}[H]
	\centering
	\caption{Quantitative error of trajectory tracking. (unit: cm)}
	\label{tab:trajectorie}
	\setlength{\tabcolsep}{6.5mm}
	\begin{tabular}{l|cc}
		\hline
		Environments & X-Axis & Y-Axis \\
		\hline
		\hline
		Basic & 3.36 & 3.58 \\
		\hline
		Occlusion & 3.61 & 3.94 \\
		\hline
	\end{tabular}
\end{table}

\subsection{Scope \& Limitations}
Experimental results demonstrate the effectiveness of RFPose-OT in basic and occlusion environments. However, RFPose-OT also has some limitations. 
1) On one hand, the operating distance of our radio system is limited to 20m. Extra transmission power would be needed to cover a larger space. 
2) On the other hand, micro hand motions may be missed by RFPose-OT for little signal power reflections.

\section{Conclusion}
In this paper, we proposed a novel RF-based 3D human pose estimation model RFPose-OT, which first transported the RF domain to the target pose domain based on OT theory, then estimated human pose keypoints from the transported RF features. To assess RFPose-OT, we conducted experiments in both basic/occlusive indoor environments and outdoor environments, and experimental results demonstrated that our proposed RFPose-OT can estimate 3D human poses with higher precision. We believe this work provides a new and valid framework to tackle RF-based human sensing tasks.

\ifCLASSOPTIONcaptionsoff
\newpage
\fi

\bibliographystyle{IEEEtran} 
\bibliography{bibsample}

\end{document}